\documentclass[a4paper]{article}

\usepackage{INTERSPEECH2020}
\usepackage[hang,flushmargin]{footmisc} 
\usepackage[hyphens]{url}
\usepackage{subcaption}
\usepackage{xcolor}
\usepackage{xspace}
\usepackage{tabularx}
\usepackage{multirow}
\usepackage{booktabs}
\usepackage[ruled, commentsnumbered]{algorithm2e}
\usepackage{enumitem}
\setlist[itemize]{leftmargin=6mm}
\usepackage{enumitem}
\usepackage{censor}
\usepackage{amsthm}

\usepackage[obeyDraft]{todonotes}

\newcommand*{\eg}{e.g.\@\xspace}
\newcommand*{\ie}{i.e.\@\xspace}

\newcommand*{\etal}{et al.\@\xspace}
\newcommand*{\spacy}{\texttt{spaCy}\xspace}
\newcommand*{\tild}{{\raise.17ex\hbox{$\scriptstyle\mathtt{\sim}$}}}
\newcommand*{\verbmobil}{\textsc{Verbmobil}\xspace}

\newcommand*{\annoPER}[1]{\colorbox{blue!30}{#1}}
\newcommand*{\annoORG}[1]{\colorbox{cyan!50}{#1}}
\newcommand*{\annoLOC}[1]{\colorbox{orange!40}{#1}}
\newcommand*{\annoTIME}[1]{\colorbox{green!20}{#1}}

\newtheorem*{definition}{Definition}

\newtheorem*{lemma}{Lemma}

\title{Privacy Guarantees for De-identifying Text Transformations}

\name{David Ifeoluwa Adelani, Ali Davody, Thomas Kleinbauer, and Dietrich Klakow
	}
\address{Spoken Language Systems Group, Saarland Informatics Campus, 
Saarland University, Germany}
	\email{\{didelani|kleiba|adavody|dietrich.klakow\}@lsv.uni-saarland.de}

\begin{document}

\maketitle
\begin{abstract}
  Machine Learning approaches to Natural Language Processing tasks
  benefit from a comprehensive collection of real-life user data. At
  the same time, there is a clear need for protecting the privacy of
  the users whose data is collected and processed. For text
  collections, such as, e.g., transcripts of voice interactions or
  patient records, replacing sensitive parts with benign alternatives
  can provide de-identification. However, how much privacy is actually
  guaranteed by such text transformations, and are the resulting texts
  still useful for machine learning?
 
  In this paper, we derive formal privacy guarantees for general text
  transformation-based de-identification methods on the basis of
  \emph{Differential Privacy}.
  
  We also measure the effect that different ways of masking private
  information in dialog transcripts have on a subsequent machine
  learning task. To this end, we formulate different masking
  strategies and compare their privacy-utility trade-offs. In
  particular, we compare a simple \textit{redact} approach with more
  sophisticated \textit{word-by-word} replacement using deep learning
  models on multiple natural language understanding tasks like named entity recognition, intent
  detection, and dialog act classification. We find that only
  word-by-word replacement is robust against performance drops in various tasks.
\end{abstract}


\section{Introduction}
\label{sec:introduction}


Machine learning approaches, in particular Deep Learning, dominate
many areas of Natural Language Processing (NLP). To reach peak
performance, they require large data sets to train models. It is thus
common to continuously collect user data after a model has been
deployed in order to augment existing training data. This practice raises a
clear need for protecting the privacy of the users whose data are
collected. For instance, commercial providers of voice assistants have
been criticized for recording and transcribing conversations of their
users\footnote{\url{https://www.bloomberg.com/news/articles/2019-04-10/is-anyone-listening-to-you-on-alexa-a-global-team-reviews-audio}}. But other domains are affected as well, for instance, patients'
health records in medical applications.


For text collections, one way to respect the users' privacy is to
sanitize each document through a de-identification process before
adding it to a data collection. De-identification requires to either
delete all sensitive information in a text, or to replace it with
benign surrogates. \todo{Insert citations.} Arguably, strict deletion
is the less desirable option because without an indication of where
the edit was made, texts can become impossible to understand or,
perhaps worse, change their meaning. To illustrate, consider the
following example from the medical domain, where the names of specific
medications are considered sensitive information:
\begin{enumerate}[label=(\arabic*), leftmargin=0.8cm]
\item Besides \textit{warfarin}, the patient is not taking any medication.
\vspace{-1mm} 
\item Besides, the patient is not taking any medication .
\end{enumerate}
A more commonly used alternative to deletion is \textit{redaction}, 
where relevant text portions are blackened rather than deleted, but
this can still impact readability. For instance, when dates, times, and
locations are considered sensitive with respect to a person's
whereabouts, a sentence such as \ref{item:ex:redacted_before} would be
rather useless in a text corpus in its redacted form
\ref{item:ex:redacted_after}:
\begin{enumerate}[label=(\arabic*), leftmargin=0.8cm]
\setcounter{enumi}{2}
\item \label{item:ex:redacted_before} How about \textit{Rick's
    Caf\'{e}} around \textit{noon} on \textit{the 15th} ?
\item \label{item:ex:redacted_after} How about \censor{Rick's
    Caf\'{e}} around \censor{noon} on \censor{the 15th} ?
\end{enumerate}
This example illustrates that de-identification can imply a trade-off
between \emph{privacy} and \emph{utility}. 
%
In order to gauge the latter, Tang \etal measure the impact of three
alternative methods for masking sensitive words on a subsequent
machine learning task \cite{Tang2004PreservingPI}. The methods consist
of two different ways of replacing words with other, randomly selected
words from the same category, and of a method for replacing words with
a specific category marker. Applying these strategies to the previous
example could lead \eg to the sentences in \ref{item:ex:vd} and
\ref{item:ex:vcm} respectively:
\begin{enumerate}[label=(\arabic*), leftmargin=0.8cm]
\setcounter{enumi}{4}
\item \label{item:ex:vd} How about \textit{London} around \textit{4
    o'clock} on \textit{the 3rd of May} ? 
\item \label{item:ex:vcm} How about \textit{$<$\footnotesize{LOCATION}$>$} around
  \textit{$<$\footnotesize{TIME}$>$} on \textit{$<$\footnotesize{DATE}$>$}? 
\end{enumerate}
\vspace{-1mm}
However, Tang \etal~\cite{Tang2004PreservingPI} do not give any formal privacy guarantees for
their methods, making it difficult to judge the privacy-utility
trade-off.

In this paper, we fill this gap by deriving formal privacy guarantees
for general text transformation-based de-identification methods on the
basis of \emph{Differential Privacy}, a well-established framework for
quantifying privacy leakage \cite{DworkC+14}. In addition, we show the
impact of five different text transformation strategies on three
common NLP tasks, when the transformed texts are used as training data
for machine learning approaches. Unlike Tang \etal, we perform our
experiments on six different corpora to gain more balanced evidence. 
To encourage reproducibility of our results, we release the code, data and models on Github\footnote{\url{https://github.com/uds-lsv/privacy-preserving-text-transformer}}

\section{Related Work}
\label{sec:related-work}

Text de-identification, also known as sanitization, is
well established in highly sensitive domains, such as \eg, for
patient health records \cite{UzunerO+07}. A large number of
de-identification methods have been suggested in that area in the
past \eg
\cite{MeystreSM+10:de-identification-review,DernoncourtF+17,KhinK+18}. While such methods are important for a number of domains, we focus here on the privacy issues associated with cloud-based dialog systems, such as \eg \cite{ChenH+17}, which are generally acknowledged (e.g. [8]) but have not yet received wide-spread attention.

In contrast, the necessity for protecting private information has long
been realized in the data mining community \cite{AggarwalCC+08}. Our
task is, however, quite different from data mining. For instance,
data mining transformations oftentimes pay special attention to
the preservation of certain statistical properties of the underlying
data, which is not a primary concern in our work, thus allowing us to
explore simpler approaches.

Probably the closest work to ours is \cite{Tang2004PreservingPI} where
the impact of data sanitization has been investigated on Automatic Speech Recognition (ASR) and call
classification tasks. It has been shown there that the spoken dialog
system trained on sanitized data achieves a comparable accuracy. In
contrast, we apply text replacement to Natural Language Understanding (NLU) tasks and show that some
replacement strategies can have a large destructive effect on the
performance of the model. We verify this by applying a
state-of-the-art deep learning model to train the NLU tasks by
fine-tuning BERT~\cite{Bert_jacob} embeddings on three tasks that include
named entity recognition, intent detection and dialog act
classification. 

Recently, Carrell \etal~\cite{parrot_attack} proposed
an attack to leak sensitive information in a transformed text,
however, this attack only works on a small dataset. The attack does
not scale to large datasets because it requires the attacker to
perform annotation of private tokens, which is costly and tedious.

\section{Privacy}
\label{sec:privacy}

A general framework for protecting privacy is \textit{Differential
  Privacy} (DP) introduced in \cite{Dwork}. DP quantifies to what
extent privacy in statistical queries is preserved while extracting
useful information from a dataset and has received increasing
attention recently as a rigorous privacy methodology. In this section,
we clarify the connection between DP and text replacement methods but
first provide some technical background on general DP.

Let $\mathcal{D}$ be the set of all possible datasets for a given
domain of data points. A key concept in DP is neighboring datasets. We
call two datasets $D_1, D_2 \in \mathcal{D}$ neighboring if they are
the same except for one data point. For example, $D_1$ and $D_2$ could
be two text corpora which differ only in one single word. The intuition
behind differential privacy, as defined below, is a guarantee that a
randomized algorithm behaves similarly on similar input datasets to a
point where the output of the algorithm does not allow to infer which
dataset was used with any relevant degree of certainty. Therefore, an
attacker cannot tell whether the aforementioned data point is
contained in the algorithm's dataset or not.

\begin{definition}{(Differential Privacy).}
  A randomized algorithm $\mathcal{M}$ is $(\varepsilon, \delta)$
  private with domain $\mathcal{D}$ if \, for all measurable sets
  $S \in \text{Range}(\mathcal{M})$ and for all neighboring datasets
  $D_1$ and $D_2$ differing in at most one data point, we have
  \vspace{-1mm}
\begin{equation}
    \Pr[\mathcal{M}(D_1) \in S] \leq \exp{(\varepsilon)} \Pr[\mathcal{M}(D_2) \in S] + \delta
    \vspace{-1mm}
\end{equation}
  
\end{definition}

\noindent
Intuitively, a $(\varepsilon, \delta)$ differential private mechanism
guarantees that the absolute value of privacy leakage will be bounded
by $\varepsilon$ with probability at least $1-\delta$ for adjacent
datasets. The higher the value of $\varepsilon$, the higher the chance
of data re-identification.

\smallskip

\vspace{-3mm}
\renewcommand{\thealgocf}{} 
\begin{algorithm}
\SetAlgoLined
\KwIn{dataset $\mathcal{D}$, token replacement policy $\pi$,
  probability parameter $p$.}

\For{$t'$ in sensitive data}
{
$r \sim  U(0, 1)$
\If{$r\leq p$}
{replace $t'$  with $t \sim \pi(t|t')$}
}
 \caption{Probabilistic Text De-identification}
\end{algorithm}
\vspace{-2mm}

To define a general algorithm for de-identification, 
let $\mathcal{T}$ denote the vocabulary of private tokens and consider
a token replacement policy
$\pi: \mathcal{T} \longrightarrow \mathcal{T}$, where $\pi(t|t')$ is
the probability of replacing $t'$ in the original text with $t$. 
We introduce a parameter $p$ to model the probability that a token
gets replaced:

\begin{lemma}
\vspace{-1mm}
  If token replacement policy $\pi$ in the algorithm is independent
  of the token to replace, \ie $\pi(t|t')=\pi(t)$, the algorithm is
  $(\varepsilon, 0)$ differentially private with:
  \vspace{-1mm}
\begin{equation}
    \varepsilon = \max_t \log\frac{1-p+p \,\pi(t)}{p \, \pi(t)}.
    \label{eqn:lemmaEqn}
    \vspace{-1mm}
\end{equation}
\end{lemma}

To prove it, we consider two neighboring datasets $D_1$ and $D_2$, which are the same except in one token. In other words, $D_2$ can be
obtained from $D_1$ by replacing a token $t_1$ in $D_1$ with $t_2$.
Using this notation, we may compute the privacy loss as follows. Let
\begin{equation}
    \varepsilon = \log \frac{\Pr[t\in \mathcal{M}(D_1)]}{\Pr[t\in \mathcal{M}(D_2)]}
\end{equation}
where $\mathcal{M}(D)$ is the dataset obtained by applying the
de-identification algorithm to the original dataset $D$, and $t$ is the
observed token in the resulting text. If $t_1$ and $t_2$ are not equal
to $t$ (i.e., $t_1$ and $t_2$ are replaced by $t$), we have :
\begin{equation}
  \label{eq:case1}
     \frac{\Pr[t\in \mathcal{M}(D_1)]}{\Pr[t\in \mathcal{M}(D_2)]}
    =\frac{p\, \pi(t|t_1)}{p\, \pi(t|t_2)} = 1
\end{equation}
where we have used this fact that replacement policies are independent
of the original tokens. On the other hand, if $t$ is equal to $t_1$,
we arrive at the following expression for the privacy loss:

\begin{equation}
  \label{eq:case2}
 \frac{\Pr[t\in \mathcal{M}(D_1)]}{\Pr[t\in \mathcal{M}(D_2)]}
    =\frac{1-p+p\, \pi(t)}{p\,\pi(t)}.
\end{equation}

We get the inverse of this expression in the opposite case $t=t_2$.
The overall privacy bound is given by the maximum of \eqref{eq:case1}
and \eqref{eq:case2} over the private tokens as stated in
\eqref{eqn:lemmaEqn}.
\qed

\medskip

The algorithm is a variant of \textit{randomized response} \cite{RR}
whose connection to differential privacy has been studied before (\eg
\cite{DworkC+14,RR,WangY+16}), although not in the context of text
de-identification. The probability parameter $p$ gives data curators
fine-grained control over the privacy-utility trade-off: an
\textit{ideal} text replacement, corresponding to $p=1$, has zero
privacy loss ($\epsilon=0$) but in cases where the replacement noise
harms the performance of models trained on the resulting data too
much, the curator might choose to use a lower probability $p$ if reduced
privacy is deemed acceptable. Our lemma allows to quantify this effect
and compare different de-identification options.

In practice, $p=1$ cannot be achieved very easily if the sensitive
tokens are identified automatically as in \eg
\cite{Tang2004PreservingPI}'s and our own experiments below. Instead,
the \textit{recall} value of the employed identification method
defines an upper bound for $p$, \eg, a recall value of $0.8$
implies that an expected $20\%$ of the sensitive tokens will not be
replaced. As $p$ approaches $0$, the $\epsilon$ value approaches
infinity, meaning that no privacy is provided.

In order to interpret our result, we consider the case where an
attacker gets hold of the fully transformed data set. The level of
privacy expressed by the lemma 
refer to the possibility of
reversing the replacement in order to reconstruct the source from a
transformed sentence, which is difficult when the privacy loss is small. 
Context information might be helpful, but in general,
original tokens can only be guessed according to their prior
probabilities which we assume to be uniform in this paper. However, the algorithm allows for certain sentences to
appear in the output untransformed, either because of the value of the
randomized response value $r$, or when the randomly chosen replacement
token happens to be identical to the source token. The privacy
guarantee given by our lemma arises from the fact that transformed and
untransformed sentences are not obviously distinguishable. In fact,
the DP parameter, $\epsilon$ can be seen as a measure of the certainty with
which an attacker can judge whether a sentence from the output was
actually part of the source text.


Referring to a specific instance of the above algorithm, \ie a fixed
choice for $p$ and $\pi$ (called a text replacement \emph{strategy}), with
tokens being either single words or multi-word expressions, we examine some straight-forward replacement strategies and the level of privacy they present in the light of our results.
\begin{description}[style=unboxed,leftmargin=0cm]
\item[\textbf{Redact}] Here, the private tokens are replaced with a
  non-word placeholder that is typically not part of the vocabulary of
  the source text \eg
  {\setlength{\fboxsep}{2pt}\colorbox{lightgray}{\texttt{IIIII}}}.
  Hence, we only fall into case \eqref{eq:case1} above, implying $\epsilon=0$
   under the interpretation outlined above: an attacker can
  decide with certainty which of the tokens were part of
  the original text but cannot infer the replaced tokens. 
\item[\textbf{Typed placeholder} (aka \textbf{value-class
    membership}~\cite{Tang2004PreservingPI})] This is akin to using private category markers like LOCATION as the replacement token. This is a strategy
  similar to redaction, providing the same level of privacy.
  However, it provides additional information about a replaced token's
  category and might thus be more useful than redaction for certain
  NLP tasks.
\item[\textbf{Named placeholder}] A fixed category exemplar is used to
  replace all private tokens of that
  category~\cite{pestian-etal-2007-shared_named_Placeholder}, e.g, all
  locations are replaced by ``London''. This strategy makes it
  slightly more difficult to judge which sentence was transformed and
  which was not, \ie $\epsilon > 0$. But for all instances that differ
  from the exemplar, it is clear that they must have been part of the
  source.
\item[\textbf{Word-by-word replacement}] We can distinguish between
  \textit{value distortion}~\cite{Tang2004PreservingPI} if the
  replacement tokens are from an external source, and \textit{value
    dissociation}~\cite{Tang2004PreservingPI} when the surrogate
  tokens are from the same corpus. The latter keeps the distribution
  of tokens in the resulting document unchanged, which might be
  relevant for some tasks. Both variants make it hard to identify
  untransformed sentences, which is reflected in lower $\epsilon$
  values.
\item[\textbf{Full entity replacement}] Text coherence could be
  improved if source tokens were consistently replaced by the same
  surrogates. However, this case is not supported by our lemma where
  we require $\pi(t|t')=\pi(t)$. Another downside of the word-by-word
  strategy is that multi-word expressions could lead to nonsensical
  replacements, \eg ``Frankfurt Airport'' could be transformed to
  ``New Francisco''. A variant is thus to replace full entities
  instead of single words. In terms of what can be captured by our lemma, this does not lead to more privacy, 
  but the expected
  gain in coherence might benefit downstream tasks.
\end{description}

\begin{table}[t]
  \centering
  \caption{Examples of the replacement strategies, using color codes
    for \annoPER{PER}, \annoLOC{LOC}, \annoORG{ORG}, and \annoTIME{TIME}.
    \vspace{-2mm}}
  \label{tab:transformation_example}
  \scalebox{0.7}{ 
  \def\arraystretch{1.3}
  \begin{tabularx}{0.65\textwidth}{|p{.15\textwidth}X|} 
    \hline
    \textbf{Replacement strategy} & \textbf{Transformed text} \\
    \hline
    No Replacement & Hi Mister \annoPER{Miller}, the                      \annoORG{Lufthansa} flight from \annoLOC{Frankfurt Airport} to \annoLOC{Rome} is leaving by \annoTIME{six pm} \\
    \hline
    Redact & Hi Mister \annoPER{IIIII}, the                      \annoORG{IIIII} flight from \annoLOC{IIIII} to \annoLOC{IIIII}  is leaving by \annoTIME{IIIII} \\
    \hline
    Typed-Placeholder & Hi Mister \annoPER{PER }, the                      \annoORG{ORG} flight from \annoLOC{LOC} to \annoLOC{LOC} is leaving by \annoTIME{TIME} \\
    \hline
    Named-Placeholder & Hi Mister \annoPER{Smith}, the                      \annoORG{SAP} flight from \annoLOC{London} to \annoLOC{London} is leaving by \annoTIME{afternoon} \\
    \hline
    Word by word & Hi Mister \annoPER{John}, the                      \annoORG{BOSCH} flight from \annoLOC{New Boston} to \annoLOC{Berlin} is leaving by \annoTIME{eleven morning} \\
    \hline
    Full entity & Hi Mister \annoPER{John},  the                      \annoORG{BOSCH} flight from \annoLOC{New York} to \annoLOC{Berlin} is leaving by \annoTIME{twelve pm} \\
    \hline 
  \end{tabularx}
}
\vspace{-4mm}
\end{table}

\begin{table*}[t]
  \centering
  \caption{Evaluation of the different token replacement strategies on
    the 6 datasets comprising of 3 tasks: NER, ID, DAC. Average performance computed from ten runs. The best Accuracy/F1-score in
    each class/task are in \textbf{bold} and the best text
    transformation result have asterisk (*). The replacement
    strategies use ground-truth annotations for the identification of
    sensitive tokens, \ie $p=1, \varepsilon=0$.\vspace{-2mm}}
  \label{tab:transf_result}
    \centering
    \footnotesize
    \scalebox{0.90}{
    \begin{tabular}{lrrrrrr}
    \toprule
     \textbf{Replacement strategy} & \multirow{2}{20mm}[0.3em] {\textbf{VerbMobil NER F1-score}} & \multirow{2}{15mm}[0.3em] {\textbf{ATIS ID Accuracy}} & \multirow{2}{17mm}[0.3em] {\textbf{SNIPS ID Accuracy}} & \multirow{2}{20mm}[0.3em] {\textbf{en-TOD ID Accuracy}}& \multirow{2}{20mm}[0.3em] {\textbf{Restaurant DAC Accuracy}} & \multirow{2}{15mm}[0.3em] {\textbf{Taxi DAC Accuracy}} \\
     \addlinespace[1.0em]
    \midrule
      No replacement & $\mathbf{88.3 \pm 0.2}$ & $\mathbf{98.4 \pm 0.2}$ & $\mathbf{98.0 \pm 0.2}$ & $\mathbf{99.4 \pm 0.0}$ &  $\mathbf{78.9 \pm 0.1}$ & $\mathbf{90.0 \pm 0.1}$ \\ \addlinespace[0.4em]
      Redact & $0.2 \pm 0.2$ & $94.8 \pm 0.2$ & $89.7 \pm 0.8$ & $97.4 \pm 0.6$ &  $75.9 \pm 0.3$ & $88.1 \pm 0.2$ \\ \addlinespace[0.4em]
      Typed-Placeholder & $0.0 \pm 0.0$  & $95.7 \pm 0.3$ & $54.1 \pm 3.8$ & $97.2 \pm 0.7$ &  $76.5 \pm 0.2$ & $87.9 \pm 0.5$ \\ \addlinespace[0.4em]
     Named Placeholder & $13.5 \pm 1.4$  & $95.9 \pm 0.3$ & $76.2 \pm 2.9$ & $98.2 \pm 0.1$ &  $77.3 \pm 0.2$ & $89.3 \pm 0.1$ \\ \addlinespace[0.4em]
     Word-by-Word & $72.6 \pm 0.3$ & $\mathbf{98.6 \pm 0.2}^*$ & ${97.5 \pm 0.3}^*$ & ${99.2 \pm 0.1}^*$ &  $78.4 \pm 0.2$ & ${89.9 \pm 0.2}^*$ \\ \addlinespace[0.4em]
     Full Entity & ${85.9 \pm 0.3}^*$  & $\mathbf{98.5 \pm 0.2}^*$ & ${97.4 \pm 0.3}^*$ & ${99.2 \pm 0.1}^*$ & ${78.5 \pm 0.1}^*$ & ${89.9 \pm 0.1}^*$ \\\addlinespace[0.4em]
      \bottomrule
    \end{tabular}
    }
    \label{Tab:result}
    \vspace{-4mm}
\end{table*}

\noindent
An example for each of these replacement strategies is given in
Table~\ref{tab:transformation_example}. Besides discussing privacy
aspects, we have speculated on the differences of the strategies on
subsequent applications. In order to verify these considerations, we
now measure the impact of the different replacement strategies
empirically.


\section{Utility}
\label{sec:utility}

We experiment with three common NLP tasks, Named Entity Recognition
(NER), Intent Detection (ID), and Dialog Act Classification (DAC),
across six different datasets (see Table~\ref{tab:datasets}). The
variety in datasets is important since what is considered sensitive
information is typically domain-dependent. Here, we consider as
private: (1) the identity of one or both speakers, (2) organizations, such as e.g., company names, etc. (3) The locations or
addresses (4) The dates and times. This private information coincides with
typical \textit{named entities (NEs)} and \textit{slot classes} in
dialog datasets such as PER (personal names), ORG (organization), LOC
(location), DATE and TIME.

\vspace{-2mm}

\subsection{Datasets}
\label{sec:datasets}

\begin{table}[tb]
  \setlength{\tabcolsep}{0.5em}
  \renewcommand{\arraystretch}{1.2}
  \centering
  \caption{Dataset summary for three different tasks: Named Entity
    Recognition (NER), Intent Detection (ID), and Dialog Act
    Classification (DAC)\vspace{-2mm}}
  \label{tab:transf_result}
    \centering
    \footnotesize
    \begin{tabularx}{\columnwidth}{lclcr@{\hskip .7mm}c@{\hskip .7mm}r@{\hskip .7mm}c@{\hskip .7mm}r}
    \toprule
      \textbf{Dataset} & \textbf{Task} & \textbf{Private} &
      \textbf{Classes} & \textbf{Train} & / & \textbf{Val.} & / &
      \textbf{Test} \\ \addlinespace[-0.25em]
      & & \textbf{Tokens} & & \multicolumn{5}{l}{\textbf{Sentences}} \\
    \midrule
      VerbMobil & NER & 5 NEs & 6 & 19K &\textcolor{gray}{/}& 2848 &\textcolor{gray}{/}& 5230  \\
      ATIS & ID & 21 slots & 21 & 4478 &\textcolor{gray}{/}& 500 &\textcolor{gray}{/}& 893  \\
      SNIPS & ID & 39 slots & 7 & 13K &\textcolor{gray}{/}& 700 &\textcolor{gray}{/}& 700  \\
      FB en-TOD & ID & 15 slots & 12 & 30K &\textcolor{gray}{/}& 4181 &\textcolor{gray}{/}& 8621  \\
      MS Restaurant & DAC & 21 slots & 24 & 20K &\textcolor{gray}{/}& 2936 &\textcolor{gray}{/}& 5859  \\
      MS Taxi & DAC & 10 slots & 18 & 16K &\textcolor{gray}{/}& 2273 &\textcolor{gray}{/}& 4597  \\
      \bottomrule
      \end{tabularx}
      \label{tab:datasets}
      \vspace{-4mm}
\end{table}

The \textbf{\verbmobil} corpus is a large collection of spontaneous
telephone conversations \cite{WeilhammerKURFS02}. In each conversation, two
speakers negotiate the details of a business meeting. The corpus
contains English, German, and Japanese conversations, however, we only
use English portion of the corpus for our experiments. The \verbmobil corpus does
not come pre-annotated with NE classes. About 20\% of the \verbmobil
corpus was thus annotated via crowd sourcing.
The remaining
80\% of the corpus was annotated automatically using
\spacy\footnote{\url{https://spacy.io}}
and post-corrected manually.

\noindent

The \textbf{ATIS}~\cite{hemphill_atis} corpus is a popular dataset for slot filling and
intent detection tasks in the Air Travel Information Services domain.
For the text transformation experiments, we map the
provided slot labels to the aforementioned named entity categories.

\noindent

\textbf{SNIPS} is another popular benchmark dataset for slot filling 
and intent detection task by SNIPS.AI~\cite{Coucke2018SnipsVP}. The
dataset consists of seven intents from different domains  such as 
\textit{``AddToPlaylist''}, \textit{``BookRestaurant''},
\textit{``GetWeather''}, \textit{``RateBook''}.

\noindent

\textbf{FB en-TOD} is a multilingual slot and intent classification
dataset recently released by
Facebook~\cite{Schuster2019CrosslingualTL}. It consists of utterances
from three languages (English, Spanish, and Thai) and three domains
(Alarm, Reminder, and Weather). In this paper, we only use the English
dataset.

\noindent

\textbf{MS Taxi} and \textbf{MS Restaurant} are two out of three dialog
challenge datasets released by Microsoft at the SLT 2018
workshop~\cite{li2018microsoft} for taxi bookings and restaurant
reservations with 19 and 29 slot types respectively and 11 dialog acts.
The number of classes for the Taxi and Restaurant datasets are 18 and
24 respectively after removing classes with less than 40 utterances.

\vspace{-2mm}

\subsection{Experiments}
\label{sec:experiments}

For all comparison experiments, we first run a baseline experiment
using the original datasets. Then, we apply the respective privacy
strategies to the training data before fine-tuning a
BERT model for token/sentence classification. We
then compare the performance of the resulting models with the baseline
with respect to the (untransformed) test set. The BERT classification
model involves \textit{fine-tuning} the pre-trained BERT embeddings on
the training data with an additional linear layer whose weights are
randomly initialized. We trained all the parameters of the model
end-to-end, including the linear layer.

For the implementation, we fine-tuned BERT on the various tasks using
the
\textit{simpletransformers\footnote{{https://github.com/ThilinaRajapakse/simpletransformers}}}
based on the \textit{transformers} library of
HuggingFace~\cite{Wolf2019HuggingFacesTS}. The hyper-parameters of the
model are 768-dimensional embedding layer (for bert-base-cased model),
batch size of 8 for NER and 16 for other tasks, maximum learning rate
of $0.00005$, maximum sequence length is 128 for NER and 64 for other
tasks. The maximum number of epochs for all experiments is 3.

\begin{figure}[tbh]
  \centering
  \vspace{-3mm}
  \includegraphics[draft=false,width=\columnwidth]{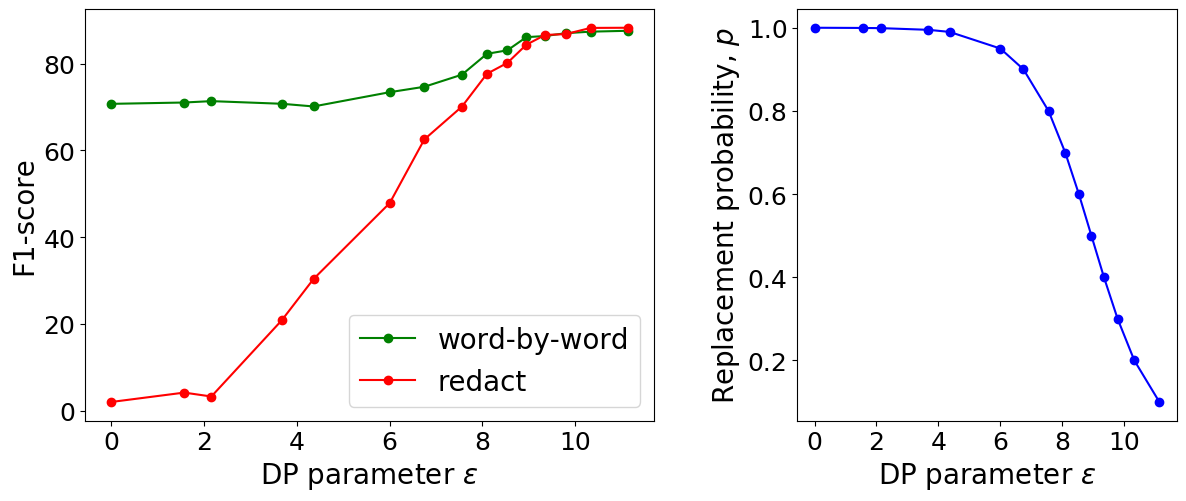}
  \vspace{-4mm}
  \caption{Connection between DP (privacy), Replacement probability $p$, and F1-score (performance) on \verbmobil.}
  \label{fig:dp_result}
   \vspace{-6mm}
\end{figure}

\subsection{Results}
\label{sec:results}

We show that the performance of the redact and word-by-word replacement strategies
can be improved by tuning the parameter $p$ described in
Section~\ref{sec:privacy}. For example, by setting $p=0.9$ (\ie
$\epsilon=6.75$), we can improve the F1-score by around $4\%$ for word-by-word and $60\%$ for redact as shown
in Figure~\ref{fig:dp_result}, demonstrating the ability to control the privacy-utility trade-off. In the word-by-word replacement
experiments, we replace a NE word $t'$ by another word $t$ of the same
entity class based on their relative frequency distribution $\pi(t)$
in the corpus.

The baseline to which we compare all other experiments is simply
trained on the original training set, \ie, without removing any
private information. On the test set, the resulting model yields a
prediction F1-score/Accuracy of $88.3\%$ (NER), $98.4\%$ (ATIS intent)
and $98.0\%$ (SNIPS intent), $99.4\%$ (en-TOD intent), $78.9\%$
(Restaurant DAC) and $90.0\%$ (Taxi DAC).

Table~\ref{Tab:result} shows the result for the different text
transformation strategies. Replacing private tokens using
\textit{redact}, \textit{typed placeholder} and \textit{named
  placeholder} strategies generally gave a worse result than the
word-by-word replacement. For NER, we observe a substantial drop in performance for redact
and placeholder approaches because the model overfits on the
replacement tokens which are expected to be absent in the test set. On the other
hand, the drop is minimal for intent and dialog act classification
tasks around ($2 - 4\%$) similar to the
observation in \cite{Tang2004PreservingPI}, except for the SNIPS
dataset with much larger reduction in performance of
$8 - 44\%$ depending on the placeholder strategy. This shows that
these transformation strategies are generally not suitable for
training NLU systems.

For the \textit{word-by-word} replacement, we observe a drop of 15\%
in F1-score when we replace all words labeled as named entities with
tokens of the same-type.
For NER, we find
that ``TIME'', ``ORG'' and ``DATE'' are most affected by the
word-by-word replacement in terms of drop in F1-score because many of
them are multi-word expressions. Thus, the three named entities gain
the most by full-entity replacement. On the other-hand, the drop is
very small ($<1\%$) for other intent and dialog act classification.

Table~\ref{tab:transformation_example} illustrates an example of the
full-entity replacement (e.g ``Frankfurt Airport'' is replaced by
``New York''). This approach gives the best performance out of all the
transformation strategies with only $2.4\%$ drop for NER. Interestingly,
there is no significant difference between its performance and the
baseline on the intent and dialog act classification tasks
across the datasets. In summary, the text obtained using the word-by-word or
full-entity text transformation are more suitable for training NLU
systems 
while protecting the
privacy of users.

\section{Conclusion}
\label{sec:discussion}

Replacing sensitive tokens with benign alternatives is a common method
for de-identifying text documents. We prove that privacy guarantees
for a formalized version of this process can be expressed in terms of
Differential Privacy. Our approach includes two parameters, $p$ and
$\pi$ that allow different replacement strategies to be expressed as
instances of the same algorithm. The respective DP-$\epsilon$ value follows
from the choices for $p$ and $\pi$, permitting a comparison of
different replacement strategies with respect to their privacy
implications.

User privacy is juxtaposed by the performance impact that a text
transformation has on subsequent machine learning tasks. We experiment
with three different NLP tasks across six different datasets and find
that both word-by-word and full entity replacement strategies are
robust against performance drops across all examined tasks.

\todo{future work}

\section{Acknowledgments}
This research has received funding by the European Union's Horizon 2020 research and 		innovation programme under grant agreement No. 3081705 -- COMPRISE (\texttt{http://www.compriseh2020.eu/}). We thank Xiaoyu Shen, Volha Petukhova, and the anonymous reviewers for their insightful comments that helped us to improve
this paper.

\bibliographystyle{IEEEtran}
\bibliography{reference}

\end{document}